\pdfoutput=1

\documentclass[11pt]{article}

\usepackage[]{EMNLP2023}

\usepackage{times}
\usepackage{latexsym}
\usepackage{bm}
\usepackage{float}
\usepackage[T1]{fontenc}

\usepackage[utf8]{inputenc}
\usepackage{multirow,tabularx}
\usepackage{arabtex}

\usepackage{makecell}
\usepackage{utf8}
\usepackage{amsmath}
\usepackage{graphicx}
\usepackage{wrapfig}
\usepackage[super]{nth}
\graphicspath{ {./images/} }

\usepackage[nopatch=eqnum]{microtype}

\usepackage{inconsolata}

\usepackage{xspace}
\usepackage{abstract}

\newcommand{\Ar}[1]{{\small \<#1>\xspace}}

\newcommand{\TrAr}[1]{\arabtrue\transfalse {\scriptsize \Ar{#1}} /\arabfalse\transtrue \RL{#1}\arabtrue\transfalse} 
\title{ArBanking77: Intent Detection Neural Model and a New Dataset in Modern and Dialectical Arabic}

\author{Mustafa Jarrar \\
  Birzeit University \\ Birzeit, Palestine \\
  \texttt{mjarrar@birzeit.edu} \And
Ahmet Birim \\
  Sestek  \\ Istanbul, Türkiye \\ \texttt{ahmet.birim@sestek.com} \And
Mohammed Khalilia \\
  Birzeit University \\ Birzeit, Palestine \\
  \texttt{ mkhalilia@birzeit.edu} \AND 
Mustafa Erden \\
   Sestek  \\ Istanbul, Türkiye \\\texttt{mustafa.erden@sestek.com} \And
Sana Ghanem \\
  Birzeit University \\ Birzeit, Palestine \\
  \texttt{ swghanem@birzeit.edu} }

\vspace{3em}
\begin{document}

\setcode{utf8}
\maketitle
\begin{abstract}

This paper presents the ArBanking77, a large Arabic dataset for intent detection in the banking domain. Our dataset was arabized and localized from the original English Banking77 dataset, which consists of 13,083 queries to ArBanking77 dataset with 31,404 queries in both Modern Standard Arabic (MSA) and Palestinian dialect, with each query classified into one of the 77 classes (intents). Furthermore, we present a neural model, based on AraBERT, fine-tuned on ArBanking77, which achieved an F1-score of 0.9209 and 0.8995 on MSA and Palestinian dialect, respectively. We performed extensive experimentation in which we simulated low-resource settings, where the model is trained on a subset of the data and augmented with noisy queries to simulate colloquial terms, mistakes and misspellings found in real NLP systems, especially live chat queries. The data and the models are publicly available at \url{https://sina.birzeit.edu/arbanking77}.
\end{abstract}

\section{Introduction}

\begin{figure}[h]
\centering
\includegraphics[width=0.45\textwidth]{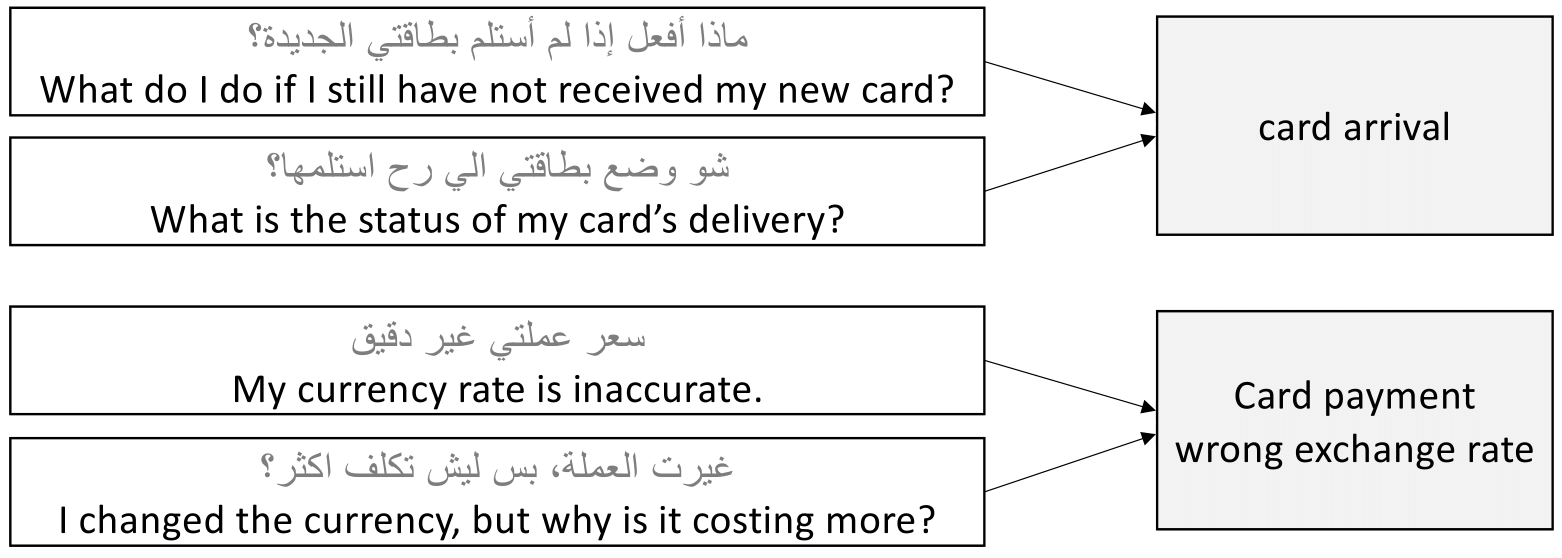}
\caption{Examples queries and their  intent.}
\label{fig:example}
\vspace*{-4mm}
\end{figure}

Intent detection falls under natural language understanding (NLU) and it aims at parsing the semantics of the user input in order to generate the best response. Intent representation is a mapping between the user request and the actions the chatbot triggers \cite{adamopoulou}. Intent detection is typically considered a classification task, where each utterance is associated with one, and sometimes multiple, intents (Figure \ref{fig:example}). 

Intent detection can be a challenging problem. The utterances during the chat are usually short, providing only a brief context to rely on when predicting the intent and the label space can be very large requiring massive data annotation. In this paper, we present an Arabic intent dataset and a Bidirectional Encoder Representations from Transformers (BERT) based intent detection model.

The Arabic corpus presented in this paper is based on the Banking77, an English question-intent corpus for banking \cite{casanueva2020efficient}. Banking77 includes 13,083 queries, each query classified into one of the 77 intents. We first arabized the English Banking77 by providing an MSA version to each of the 13,083 queries, resulting in 15,537 MSA queries (some queries have more than one MSA variation). The arabization was done semi-automatically, first we used Google Translate and then manually verified and revised each query. Second, each query was manually re-written in the Palestinian dialect, resulting in 15,867 queries, which makes the data linguistically more representative from various aspects including phonology, morphology, lexicon, and syntax \cite{EJHZ22,JHRAZ17}. The final dataset contains 31,404 queries, which was used to train a BERT-based model on intent detection task. 

The rest of the paper is organized as follows: section \ref{sec:related_work} reviews the related work
, section \ref{sec:corpus} presents the ArBanking77 corpus including data arabization and localization, section \ref{sec:model} presents the model architecture and training, section \ref{sec:results} presents the results for intent detection, section \ref{sec:conclusion} presents our conclusion and section \ref{sec:limitations} states limitations.

\section{Related Work}
\label{sec:related_work}
Arabic has a limited number of available labeled datasets, especially for dialectal and domain-specific tasks \cite{DH21,KAJ21}. Due to data scarcity in Arabic language, research on Arabic intent detection is almost non-existent. Others have also stated the same, where conversational machine learning systems in Arabic are limited due to deficiency of datasets \cite{9723043} and Arabic conversational systems are lagging behind in applying the latest technology \cite{ahmed2022arabic}.

One of the closest work to Arabic intent detection is purposed in \cite{mental_health_intent}. The authors proposed intent detection model for the mental health domain in Tunisian Arabic. The idea is to classify the patient utterance or concern into five aspects: depression, suicide, panic disorder, social phobia and adjustment disorder. The data set was collected by simulating a real-life psychiatric interview where a 3D human avatar plays the doctor and asks the patient questions in Tunisian Arabic. The patient, in return, interacts with the avatar by answering the questions vocally, then the audio is transcribed to text. The authors used BERT as the encoder and added five binary classifiers, one classifier for each intent, achieving 0.94 F1 score.

\citealt{hijjawi2013user} classified question and non-question utterances in chatbots. Decision trees were used to perform the classification and the model was integrated into ArabChat \cite{arabchat} to classify utterances before processing them. \citet{joukhadar2019arabic} published a corpus in the Levantine Arabic dialect consisting of 873 sentences manually tagged with one of eight acts (greetings, goodbye, thanks, confirm, negate, ask/repeat, ask for alternative, and apology). The authors tried two features including Term Frequency-Inverse Document Frequency (TF-IDF) and $n$-gram. They also experimented with multiple classifiers and they concluded that Support Vector Machine (SVM) with 2-gram features performed the best at 0.86 accuracy.

\citet{elmadany2018arsas} introduced a speech-act recognition and sentiment dataset (ArSAS). About 21K tweets were collected and manually labeled with two types of classes: speech-act and sentiment. Speech-act labels include expression, assertion and question, while the sentiment labels are negative, positive, neutral and mixed.  \citet{algotiml2019arabic} trained two models on the ArSAS dataset, a Bi-directional Long-Short Term Memory (BiLSTM) and SVM and achieved an accuracy of 0.875 and a macro F1 score of 0.615. \cite{Zhou2022} proposed a contrastive based learning for out-of-domain data and tested the performance on multiple datasets including the Banking data \cite{casanueva2020efficient} and they demonstrated improvement of the out-of-domain data without sacrificing performance on in-domain-data. 

Other related languages for which intent detection was studied is Urdu. In \cite{urdu_intent}, the authors translated the Air Travel Information System (ATIS) \cite{atis} and AOL datasets from English to Urdu and performed intent detection using a combination of CNNs, LSTMs and BiLSTMs models. For ATIS, CNN performed the best at 0.924 accuracy, while for AOL, BiLSTM achieved the highest performance at 0.831 accuracy. In later work the authors improved the accuracy to reach 0.9112 \cite{improved_urdu_intent}. ATIS was also used for intent detection in the Indonesian language \cite{indonesian_intent} and the authors reported an accuracy of 0.9584 using a CNN-based model. \cite{Basu2022} utilized Snips \cite{Coucke2018} and ATIS to train a meta-learning approach with contrastive learning for intent detection and slot-filling. Snips dataset covers multiple domains including restaurants, books, weather and music, making it more challenging than ATIS. The data is collected using Snips personal assistant and contains 16K queries labeled with 7 intents.

The reader may have already noticed that we could not find relevant work related to Arabic intent detection recognition or any related work on labeled Arabic intent datasets. In this paper, we attempt to address these two issues, Arabic intent corpus and intent recognition. We present the ArBanking77, an Arabic intent dataset, which was arabized and localized from the Banking77 English dataset \cite{casanueva2020efficient}. ArBanking77 was also augmented with thousands of additional MSA and Palestinian dialect queries, resulting in a final dataset of 31,404 queries and 77 intents. ArBanking77 was used to fine-tune BERT-based model, achieving an F1-score of 0.9209 and 0.8995 on MSA and Palestinian dialect, respectively.

When deploying a fine-tuned intent detection model inside a chatbot system, other modules might be needed to better understand user queries, such as spell corrections \cite{eryani2020spelling}, named entity recognition \cite{JKG22,LJKOA23}, word-sense disambiguation \cite{HJ21b,JMHK23}, synonymy expanding \cite{GJJB23,JKKS21}. 

\section{The ArBanking77 Corpus}
\label{sec:corpus}

The ArBanking77 corpus is derived from the Banking77 dataset \cite{casanueva2020efficient} that consists of 13,083 queries and 77 classes (intents) and that is open under the (CC-BY-4.0) license. Banking77 was designed to focus on a fine-grained single domain, banking. Each query is labeled with one of the 77 classes. Example intents from the dataset include \textit{card arrival}, \textit{Personal Identification Number (PIN) blocked}, \textit{card linking}, \textit{exchange rate} and \textit{age limit}. The number of queries per class ranges between 75 to 227, with an average of 170 queries per intent. The original Banking77 dataset is divided into train and test dataset, their statistics are presented in Table \ref{tab:banking77_stats}.

\begin{table}[ht!]
\small
\centering
\begin{tabular}{|l|l|l|}
\hline
 & Train Set & Test Set \\ \hline
Query count & 10,003 &  3,080 \\ \hline
Avg word count & 11.95 &  10.95 \\ \hline
Min word count  & 2  &  2  \\ \hline
Max word count  &79  &  69 \\ \hline
Std of word count & 7.89 & 6.69 \\ \hline

\end{tabular}
\caption{Statistics of the Banking77 English dataset}
\label{tab:banking77_stats}
\end{table}

Banking77 was arabized and localized into ArBanking77 by 26 annotators through multiple phases and over several months. Each query in the Banking77 has at least two corresponding queries in the ArBanking77 (at least one query written in each MSA and Palestinian dialect). 

\subsection{Phase I: Arabization and Localization}
The first step was the translation of the Banking77 from English into MSA. We used Google Translate API to translate the 13,083 queries. For each original English query, $j$, where $0 < j < m$ and $m = 13,083$, we form the following tuple:
\begin{gather*}
(q_j^i, q_j^{En}, q_j^{MSA_1}, q_j^{MSA_2}, q_j^{PAL_1}, q_j^{PAL_2})\\  \forall 0 < j < m
\end{gather*}

\noindent where $q_j^i$ is the query's intent, $q_j^{En}$ is the original English query from Banking77, $q_j^{MSA_1}$ is the MSA translation, $q_j^{MSA_2}$ is a second MSA query, $q_j^{PAL_1}$ is the Palestinian query, and $q_j^{PAL_2}$ is a second Palestinian query.

Each annotator was asked to understand the English query and its intent, then: (i) review $q_j^{MSA_1}$, and revise it if needed; (ii) optionally write $q_j^{MSA_2}$, (iii) write a $q_j^{PAL_1}$ query, and (iv) optionally write a $q_j^{PAL_2}$ query. The annotators performed these steps according to the following arabization and localization guidelines:
\begin{itemize} 

    \item $q_j^{MSA_1}$ should be revised in case of incorrect translation. We also ensured the translation is adapted to the banking domain. For example, \textit{transfer} was incorrectly translated into \TrAr{نَقل}(ship) instead of \TrAr{تحويل} (money transfer); \textit{activate} was translated to \TrAr{تنشيط}, which is not semantically wrong, but it should be  \TrAr{تَفعيل}, as it is the common term used in the banking domain. The total number of revised translations is 2,104 ($\sim16\%$).
    
    \item $q_j^{MSA_2}$ is optionally written by the annotator if there is a need to add an extra formulation of the MSA query. For example, Personal Identification Number might be translated in $q_j^{MSA_1}$ as ({\small \<رقم التعريف الشخصي>}) and ({\small \<الرقم السري>}) as a second formulation in $q_j^{MSA_2}$.
    
    \item $q_j^{PAL_1}$ is the formulation of the query in the Palestinian dialect, reflecting the terminology Palestinians naturally use in banking services. 
    
    \item $q_j^{PAL_2}$ is optionally written by the annotator if there is a need to add an extra formulation of the query in the Palestinian dialect.
\end{itemize}

This phase was carried out by 26 annotators, who are \nth{3} and \nth{4} year college students. Each annotator was given about 500 $q_j^{En}$ queries and their translations ($q_j^{MSA_1}$) to revise. Based on $q_j^{En}$ and $q_j^{MSA_1}$, annotators also provided $q_j^{MSA_2}$, $q_j^{PAL_1}$, and $q_j^{PAL_2}$. When generating PAL queries, annotators had access to both English and MSA queries, which may bias the PAL query towards MSA. However, we verified that this is not a concern as the lexical overlap between MSA and PAL is significant (Section \ref{sec:lexical_overlap}). Furthermore, in order to diversify the queries, we avoided having all queries in one intent reviewed and written by one annotator only. Instead, each intent was divided among multiple annotators, usually 2-5 annotators.

\subsection{Phase II: Review}
\label{sec:phase2}
To control and verify the quality of the data generated in Phase I, we performed a final manual review. Each of the 26 annotators, employed for phase I, was assigned a set of queries to review. On average three intents were assigned to each reviewer and we ensured that all queries belonging to one intent are assigned to the same reviewer. In order to increase data labeling consistency, we added the constraint that classes assigned to one reviewer should be relevant to each other (i.e., card arrival, card linking, card activation). Each reviewer was asked to pay attention to the following issues: (i) The MSA and Palestinian queries should be acceptable, semantically correct and well-formulated; (ii) all queries in one intent belong to that intent, and not to other intents (labeling consistency); and (iii) spelling mistakes are ignored in order to simulate common errors and noise in real NLP systems, especially in live chat queries.

Once the review is complete, we revised duplicate queries by introducing additional variations to make them unique. Duplicate queries can arise when we have many-to-one translations, in other words, multiple English queries are translated into one Arabic query (see examples in Table \ref{tab:many_to_one_translation}).

\begin{table*}[ht]
\small
\centering
\begin{tabular}{|p{8.5cm}|p{7cm}|}
\hline
\textbf{English Queries} & \textbf{Arabic Query} \\ \hline
Can you tell me the restrictions for the disposable cards? & 
\multirow{2}{7cm}{ {\tiny \begin{arabtext}هل يمكنك إخباري بالقيود المفروضة على البطاقات التي تستخدم لمرة واحدة.\end{arabtext}}} \\
Can you please inform me of the restrictions for the disposable cards. & \\ \hline

How is an exchange rate calculated? & 
\multirow{2}{7cm} {{\tiny {\begin{arabtext}كيف يتم حساب سعر الصرف؟\end{arabtext}}}} \\
How are your exchange rates calculated? & \\ \hline
\end{tabular}
\caption{Examples of many-to-one English-Arabic translation.}
\label{tab:many_to_one_translation}
\end{table*}

Our final ArBanking77 dataset (Table \ref{tab:arbanking77_size}) consists of 31,404 queries in total, 2.4x larger than the Banking77 dataset. On average, there are 408 queries per intent (202 MSA queries/intent and 206 Palestinian queries/intent). We further divided our training data into train and validation sets, by sampling 90\% of the queries in the $i$th class to the training set and the remaining 10\% were included in the validation set. This is contrary to the train/test only split cited in \cite{casanueva2020efficient}, in which they stated small data size as the reason for not introducing a validation set.

Table \ref{tab:arbanking77_stats} presents some statistics about ArBanking77. From Table \ref{tab:arbanking77_stats} we observe that the dialectal queries are shorter than their corresponding MSA queries. In MSA the average number of words in a query is 9.85, while it is 8.06 in the Palestinian queries.
This is expected as in some cases dialectical Arabic omits interrogative nouns such as ({\small\<هل>}), so an MSA query such as ({\small \<هل يوجد شروط للعمر؟>}/are there age requirements?) is phrased in Palestinian dialect as ({\small \<في شروط علعمر؟>}). In other cases, functional words such as prepositions ({\small \<عن>/from or about, \<على>/on or above, \<إلى>/to or at, \<في>/in or into}) are used as prefixes or suffixes. For instance, the phrase ({\small \<على العمر>}) in MSA is (\TrAr{علعُمر}) in the Palestinian dialect, where ({\small \<على>}) is used as a prefix in the word (\TrAr{علعُمر}). For discussion on the orthography of Arabic dialect, see \cite{ANMFTM23,EJHZ22,JHAZ14}

\begin{table*}[ht!]
\small
\centering
\begin{tabular}{|l|c|c||l|}
\hline
 & MSA & PAL & \multirow{2}{*}{Total}  \\
 & {\small ($q_n^{MSA_1} + q_n^{MSA_2}$)} & {\small ($q_n^{PAL_1} + q_n^{PAL_2}$)} &  \\ \hline
Train & 10,733 & 10,826 & 21,559 \\ \hline
Validation & 1,230 & 1,234 & 2,464 \\ \hline
Test & 3,574 & 3,807 & 7,381 \\ \hline \hline
Total & 15,537 & 15,867 &  \textbf{31,404} \\ \hline
\end{tabular}
\caption{Size of ArBanking77}
\label{tab:arbanking77_size}
\end{table*}

\begin{table}[ht!]
\small
\centering
\begin{tabular}{|l|r|r||r|}
\hline
& MSA & PAL & Overall \\ \hline
Avg word count    & 9.85 & 8.06 & 8.95 \\ \hline
Std of word count & 6.54 & 4.66 & 5.74 \\ \hline
Min word count    & 2 & 2 & 2 \\ \hline
Max word count   & 68 & 54 & 68 \\ \hline
\end{tabular}
\caption{Statistics of ArBanking77 dataset}
\label{tab:arbanking77_stats}
\end{table}

\subsection{Lexical Relation between MSA and PAL}
\label{sec:lexical_overlap}
Arabic is a highly diglossic language, meaning that two or more distinct  languages are spoken within a given region, which is a phenomenon in the Arab countries \cite{J21}. Sometimes MSA is significantly different from colloquial dialects \cite{JZHNW23,KAJ21}, where they can be mutually unintelligible. Because of that MSA and PAL have many differences making it harder to apply MSA NLP tools to PAL. In this section, we will study the lexical difference between MSA and PAL, although the differences extend beyond lexical to include morphology, phonology, orthography, semantic and syntactic.

To measure the lexical overlap between MSA and PAL, we computed the Jaccard Index for each parallel pair (MSA and PAL) and averaged the results across the entire dataset. We found that the mean Jaccard index is 0.16, median 0.13 and standard deviation 0.13. Others have also studied the lexical overlap between MSA and PAL and reported similar results. For instance, \cite{KWAIK20182} measured the overlap between MSA and other dialects including PAL on two parallel datasets, the Parallel Arabic Dialect Corpus and Multi-Dialectal Arabic and reported Jaccard Index of 0.19 and 0.16, respectively. This shows that for diaglossic languages such as Arabic, training on one variation is not necessarily extensible. Later in section \ref{sec:zero_shot}, we will explore zero-shot learning to illustrate the effect of lexical differences on model performance.  

\section{Intent Detection Model}
\label{sec:model}
We fine-tuned a BERT-based model on an intent detection task using the ArBanking77 dataset. In this section, we will go over the model details.

\subsection{Model Architecture}
Our model is based on BERT, a transformer-based language representation for natural language processing \cite{devlin2018bert}. BERT was developed by Google in 2018 as a solution for the most common language tasks such as sentiment analysis, named entity recognition, and question answering. BERT is built using transformers, which is a deep learning architecture that solves sequence-to-sequence tasks in NLP and relies on the attention mechanism that learns the alignment between words in a given sequence. Transformers include two components: an encoder that encodes the input text and a decoder that produces a prediction for the task, such as predicting masked token or predicting next sentence. In this paper, BERT encoder is fine-tuned on Arabic intent detection task using the ArBanking77 dataset. 

For intent detection, a single linear layer was added on top of BERT transformer layers to perform the intent classification task.

\subsection{Model Training}
\label{sec:training}
We fine-tuned multiple pre-trained transformer models, which will be discussed in the next section. The hyperparameters we used are: learning rate, $1e^{-3} < \eta < 5e^{-5}$, and batch size, $B = \{16, 32, 64\}$. We ran approximately 30 experiments, with an average run-time per experiment < 2 hours, depending on model parallelism. The best performing hyperparameters were $\eta = 4e^{-5}$ and $B=64$, with maximum sequence length of 128, maximum of 20 epochs and early termination if there is no improvement on the validation data after five epochs. Model training was performed using our Nvidia Tesla P100 16GB GPU card.

\section{Experiments and Results}
\label{sec:results}
We ran multiple experiments with different models and data configurations. In section \ref{sec:zero_shot}, we evaluate zero-shot learning, section \ref{sec:other_models} benchmarks multiple pre-trained transformer models on Arabic data, section \ref{sec:low_resource_sim} simulates low-resource settings and section \ref{sec:error_sim} simulates different spelling errors that are commonly found in the Arabic language. We report the model performance on the test set using macro F1, precision and recall scores. 

When training the models on the full dataset, we used the train, validation and test split listed in Table \ref{tab:arbanking77_size}, where 21,559 queries used for training and 2,464 served as the validation set. In low-resource settings we experimented with different training and validation data sizes (Section \ref{sec:low_resource_sim}), but the test set size remained at 7,381 queries. In noise and error simulation experiments we used the same test set with 7,381 queries, but errors were injected into the test queries as we will explain in Section \ref{sec:error_sim}.

\subsection{Zero-Shot Cross-Lingual Transfer Learning}
\label{sec:zero_shot}
In some cases, zero-shot cross-lingual transfer learning can yield good results and may help us avoid the manual data annotations. In this section, we study how zero-shot cross-lingual transfer learning perform on both MSA and PAL using multi-lingual BERT (mBERT) \cite{devlin2018bert} and GigaBERT \cite{lan2020gigabert}. mBERT is trained on 104 languages including Arabic, which is based on MSA data from Wikipedia with less than 1.4 gigabytes and only 7,292 tokens \cite{app12115720}. GigaBERT was trained for Arabic NLP tasks and English-to-Arabic zero-shot transfer learning. The data contained about 13 million articles from different sources and augmented with code-switched samples to improve cross-lingual learning.

In one set of experiments we evaluated zero-shot cross-lingual transfer learning on PAL test set by fine-tuning mBERT on ArBanking77 MSA training dataset, which yielded 0.5968 F1-score (Table \ref{tab:zero_shot}). In the second set of experiments we performed zero-shot cross-lingual transfer learning on both MSA and PAL by fine-tuning GigaBERT and mBERT on the English Banking77 training data. On MSA, GigaBERT and mBERT achieved 0.5047 and 0.1774 F1-score, respectively. The performance is even lower on PAL with GigaBERT and mBERT performing at 0.3507 and 0.0903 F1-score, respectively. These experiments demonstrate the performance of multilingual pre-trained models falls behind on MSA and is significantly lower for dialectical Arabic, which begs the need for MSA and dialectical Arabic data annotations. 

\begin{table*}[ht]
\small
\centering
\begin{tabular}{|l||l||c|c|}
\hline
\textbf{Pre-trained Model} & \textbf{Training Data} & \textbf{MSA F1} & \textbf{PAL F1} \\ \hline 
Multi-lingual BERT (uncased) & ArBanking77 (MSA) & - & 0.5968\\
GigaBERT & Banking77 (English) & 0.5047	& 0.3507 \\
Multi-lingual BERT (uncased) & Banking77 (English) & 0.1774	& 0.0903 \\
\hline
\end{tabular}
\caption{Performance of zero-shot learning.}
\label{tab:zero_shot}
\end{table*}

\subsection{Pre-Trained Transformers Benchmark}
\label{sec:other_models}

As we observed in the pervious section, multilingual pre-trained transformers did not perform well on MSA and PAL. In this section, we evaluate various Arabic pre-trained transformer models in addition to mBERT on ArBanking77 dataset. We benchmark against the following models: 

\noindent \textit{AraBERT} \cite{antoun2020arabert}: trained on two major datasets, Abu El-Khair, a 1.5B words Arabic Corpus \cite{el20161} and the Open Source International Arabic News Corpus (OSIAN), which consists of 3.5 million articles (1B tokens), from 31 news sources in 24 Arab countries \cite{osian}. The final size of AraBERT dataset is 70M sentences, corresponding to about 24GB of text.

\noindent \textit{ARBERT} \cite{arbert_marbert}: trained on 61GB (6.5B tokens) of MSA text in books, news articles, Gigaword \cite{gigaword}, Open Super-large Crawled Almanach coRpus (OSCAR) \cite{oscar}, OSIAN and the Wikipedia Arabic \cite{Wikiextractor2015}.

\noindent \textit{MARBERT} \cite{arbert_marbert}: trained on dialectical Arabic collected from Twitter.

\noindent \textit{MARBERTv2} \cite{arbert_marbert}: trained on the ARBERT MSA data in addition to dialectical Arabic, has longer sequence length, trained for more epochs and contains a total of 29B tokens.

\noindent \textit{QARiB} \cite{qarib}: Qatar Computing Research Institute (QCRI) Arabic and Dialectal BERT trained on Arabic Gigaword Fourth Edition (1B words), Abu El-Khair Corpus (1.5B words) and Open Subtitles (0.5B words).

\noindent \textit{CAMeLBERT-Mix} \cite{camelbert}: trained on a mix of MSA data that includes Gigaword Fifth Edition, Abu El-Khair Corpus, OSIAN, Arabic Wikipedia, OSCAR, dialectical Arabic that covers Levantine and Gulf regions, and a subset of the OpenITI corpus \cite{nigst2020openiti}

Results for those models are presented in Table \ref{tab:other_models}, sorted by the PAL test F1-score. AraBERTv2 gives the best F1-score on both MSA and PAL with 0.9209 and 0.8995, respectively. In the remaining experiments, we will use AraBERTv2 given that it achieved the best performance. 

\begin{table*}[ht]
\small
\centering
\begin{tabular}{|l||ccc||ccc|}
\hline
         & \multicolumn{3}{c||}{\textbf{MSA Test}} & \multicolumn{3}{c|}{\textbf{PAL Test}} \\ \cline{1-7} 
\textbf{Pre-trained Model} & \textbf{Precision} & \textbf{Recall} & \textbf{F1} & \textbf{Precision} & \textbf{Recall} & \textbf{F1} \\ \cline{1-7} 
AraBERTv2 & \textbf{0.9231}	& \textbf{0.9212}	& \textbf{0.9209} & \textbf{0.9004} &	\textbf{0.9025} &	\textbf{0.8995} \\
MARBERTv2 & 0.9161	& 0.9142	& 0.9138 & 0.8983	& 0.8981	& 0.8962 \\
ARBERT & 0.9103	& 0.9121	& 0.9115 & 0.8810	& 0.8923	& 0.8899\\
QARiB & 0.9147	& 0.9123	& 0.9121 & 0.8846	& 0.8864	& 0.8835\\
CAMeLBERT-Mix & 0.9149	& 0.9133	& 0.9128 & 0.8855	& 0.8854	& 0.8830 \\
MARBERT & 0.9106	& 0.9075	& 0.9070 & 0.8817	& 0.8817	& 0.8789\\
Multi-lingual BERT & 0.8888	& 0.8872	& 0.8862  & 0.8598	& 0.8623	& 0.8578\\
\hline
\end{tabular}
\caption{Performance of various pre-trained transformers on ArBanking77}
\label{tab:other_models}
\end{table*}

Those results are based on fine-tuning the models on the manually reviewed translations. To see if the manual review of the translations improves the model performance we fine-tune two additional AraBERTv2 models. One using the original machine translated data and the second with the manually reviewed data. Note that both training datasets contain MSA only data, since Google Translate will produce MSA translation. Fine-tuning with the original translations results in F1-scores of 0.9099 and 0.7945 for MSA and PAL, respectively. When the data is manually reviewed the F1-scores are 0.9117 and 0.7918 for MSA and PAL, respectively. A very small difference, yet it was important to review the translations to adapt it to the banking domain. 

\subsection{Low-Resource Simulation}
\label{sec:low_resource_sim}
This section aims to investigate the impact of the size of the training set on the model performance. Since data labeling is typically expensive it is important to estimate the number of samples one needs to achieve good and acceptable accuracy. We conducted several experiments with different training data sizes: 20\% (of the training queries per intent were randomly sampled), 50\% and 100\% (the entire training set). Throughout all the experiments, we evaluated our model on same test set, which contains 7,381 queries.

\begin{table*}[ht]
\small
\centering
\begin{tabular}{|l||ccc||ccc|}
\hline
         & \multicolumn{3}{c||}{\textbf{MSA Test}} & \multicolumn{3}{c|}{\textbf{PAL Test}} \\ \cline{1-7} 
\textbf{\% of data} & \textbf{Precision} & \textbf{Recall} & \textbf{F1} & \textbf{Precision} & \textbf{Recall} & \textbf{F1} \\ \cline{1-7} 
20\% & 0.8825 & 0.8755 & 0.8758 & 0.8441 & 0.8403 & 0.8363 \\
50\% & 0.9117 & 0.9094 & 0.9088 & 0.8909 & 0.8903 & 0.8888 \\
100\% & 0.9231 & 0.9212 & 0.9209 & 0.9004 & 0.9025 & 0.8995 \\
\hline
\end{tabular}
\caption{Results on the ArBanking77 MSA and PAL test sets in low-resource settings}
\label{tab:low_resource_results}
\end{table*}

Results with different low-resource settings are presented in Table \ref{tab:low_resource_results}. The average increase in F1-score as we increase the training data size is about 2.26\% and 3.16\% on the MSA and PAL test datasets, respectively, which indicates the impact of the training dataset size is more noticeable on the dialectical Arabic. We also notice that the performance on the PAL test is consistently lower than MSA test. The performance gap between MSA and PAL is 2.14\%, 2\%, and 3.95\% F1-score when training with 100\%, 50\% and 20\% of the data, respectively. The largest performance gap between MSA and PAL is at the lowest setting (20\%), after that the performance gap stabilizes. Lower performance on dialectical data could be due AraBERT \cite{antoun2020arabert} not being sufficiently exposed to the Palestinian dialect during the pretraining phase. In general, dialectical Arabic is typically noisier and does not follow consistent orthography as MSA.

Surprisingly, the performance on the MSA and PAL test sets using only 20\% of the training data is impressive at 0.8758 and 0.8363 F1-scores, respectively. This indicates that we can expect to achieve an acceptable performance on other low-resource dialectical Arabic on intent detection task.

\subsection{Noise and Error Simulation}
\label{sec:error_sim}
Colloquial words, misspellings and different word variations present a challenge to chatbots. Therefore, in this section we aim to measure the robustness of our dataset and model. We experimented with three types of error and noise simulations: (1) common spelling errors (\textit{sim\textsubscript{c}}), (2) simulated errors (\textit{sim\textsubscript{s}}), and (3) keyboard-related errors (\textit{sim\textsubscript{k}}) - see Appendix \ref{app:err_sim} for the details.

We performed experiments with and without training data augmentation. In case of augmentation, train and test sets were augmented in slightly different fashion. For training, about 50\% of the queries were augmented with \textit{sim\textsubscript{s}} and the other 50\% were augmented with \textit{sim\textsubscript{k}}. The original data was combined with the augmented data resulting in 43,118 queries in the training set. We evaluated the model on three versions of the test set, one version injected \textit{sim\textsubscript{c}} errors in each query, the second version using \textit{sim\textsubscript{s}} and the third with \textit{sim\textsubscript{k}}.

Results of the combined low-resource and error simulations are summarized in Table \ref{tab:noise_simulation_results}. Due to the number of experiments, we only reported the macro F1-score. We see a similar trend to the results presented in Section \ref{sec:low_resource_sim}, the model performance on the PAL test set is consistently lower than MSA test set across all experiments. We also notice that the model is more sensitive to some errors introduced into the test set. 

We performed the experiments using two trained models, with and without training augmentation. In both models we see similar behaviour, where we observe that the average drop in performance, when reducing training set size, on PAL-\textit{sim\textsubscript{c}} across all data settings is about 3.38\%, compared to 2.37\% on MSA-\textit{sim\textsubscript{c}}. Similar pattern is also observed on the PAL-\textit{sim\textsubscript{s}} and MSA-\textit{sim\textsubscript{k}}, with an average performance drop of 3.39\% and 2.16\%, respectively. However, we see a lower performance on PAL-\textit{sim\textsubscript{s}} with an average drop in F1-score by 4.2\%, compared to 2.19\% on MSA-\textit{sim\textsubscript{s}}. From that, we learn that the model performance is stable on MSA regardless of the type of errors we inject into the data, however, on PAL we see more volatility and sensitivity in the model performance when injecting \textit{sim\textsubscript{s}} errors. Those findings reveal that BERT is more susceptible to the removal of spaces in dialectical Arabic since that results in combining two or three tokens into one. This issue is exacerbated further in dialectical Arabic since it lacks consistent orthography compared to MSA.  

\begin{figure}[h]
\centering
\includegraphics[width=0.4\textwidth]{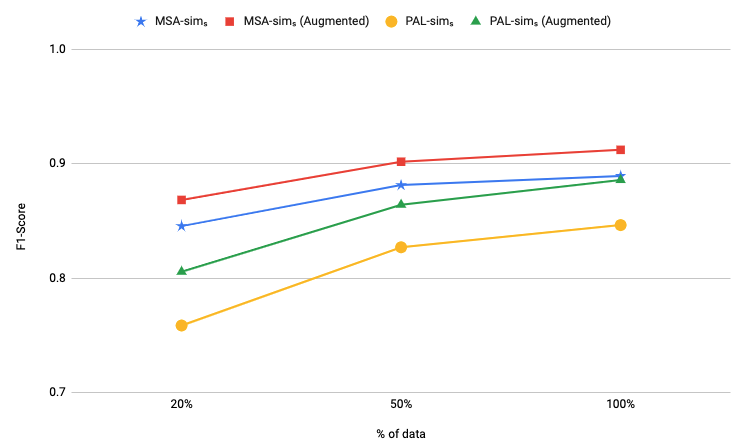}
\caption{MSA-\textit{sim\textsubscript{s}} vs. PAL-\textit{sim\textsubscript{s}} F1-scores with low-resource settings, (Augmented) indicates that the training data was augmented.}
\label{fig:20_f1_msa_pal_aug_t_f}
\end{figure}

Despite those results, we see that augmenting the training data did help close the performance gap between the PAL and MSA. Figure \ref{fig:20_f1_msa_pal_aug_t_f} zooms in a little more into the performance on MSA-\textit{sim\textsubscript{s}} and PAL-\textit{sim\textsubscript{s}} with and
without training augmentation. Three observations to make from Figure \ref{fig:20_f1_msa_pal_aug_t_f}: 1) MSA performance is better than PAL regardless of data augmentation, 2) augmenting the training data closes the performance gap between PAL-\textit{sim\textsubscript{s}} (augmented) and MSA-\textit{sim\textsubscript{s}} (without augmentation), 3) the average F1-score gain after training with augmented data on PAL-\textit{sim\textsubscript{s}} (4.12\%) is larger than MSA-\textit{sim\textsubscript{s}} (2.2\%). The improvements are less noticeable on \textit{sim\textsubscript{c}} and \textit{sim\textsubscript{k}}.

\begin{table*}[!htbp]
\small
\centering
\begin{tabular}{|c|c||ccc||ccc|}
\hline
\multirow{2}{*}{Train Augmentation} & 
\multirow{2}{*}{Test Augmentation} & 
\multicolumn{3}{c||}{\textbf{MSA Test}} & 
\multicolumn{3}{c|}{\textbf{PAL Test}} \\ \cline{3-8}
 & & \textbf{20\%} & \textbf{50\%} & \textbf{100\%} & \textbf{20\%} & \textbf{50\%} & \textbf{100\%}\\ \cline{1-8} 
\multirow{3}{*}{None} & None & 0.8758 & 0.9088 & 0.9209 & 0.8363 & 0.8888 & 0.8995 \\
& $sim_c$ & 0.8452 & 0.8795 & 0.8981 & 0.7933 & 0.8435 & 0.8637 \\
& $sim_s$ & 0.8454 & 0.8813 & 0.8893 & 0.7585 & 0.8269 & 0.8463 \\
& $sim_k$ & 0.8392 & 0.8648 & 0.8844 & 0.7942 & 0.8428 & 0.8634 \\ \cline{1-8} 
\multirow{3}{*}{$sim_s$/$sim_k$} & None & 0.8801 & 0.9126 & 0.9207 & 0.8421 & 0.8901 & 0.9018 \\
& $sim_c$ & 0.8583 & 0.8922 & 0.9001 & 0.8065 & 0.8602 & 0.8711 \\
 & $sim_s$ & 0.8683 & 0.9017 & 0.9121 & 0.8055 & 0.8641 & 0.8857 \\
& $sim_k$ & 0.8499 & 0.8833 & 0.8909 & 0.8086 & 0.8529 & 0.8749 \\
\hline
\end{tabular}
\caption{Performance in terms of F1-scores of models trained on the combined MSA and PAL datasets when simulating low-resource setting (20\% of the data) and different types of noise, "None" refers to the clean dataset while the percentages in the header indicate the percentage of training data used.}
\label{tab:noise_simulation_results}    
\end{table*}

\begin{figure}[h]
\centering
\includegraphics[width=0.4\textwidth]{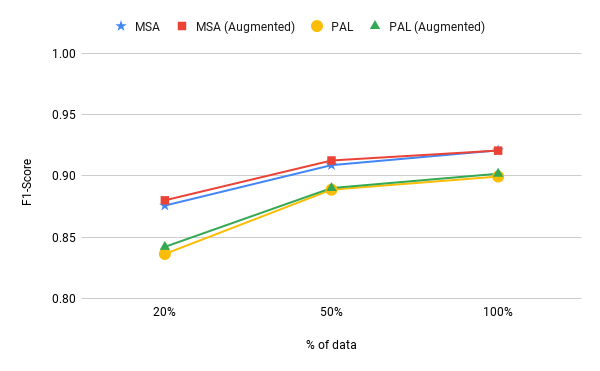}
\caption{MSA vs. PAL clean sets F1-scores with low-resource settings and data augmentation, (Augmented) indicates that the training data was augmented.}
\label{fig:f1_msa_pal_aug_t_f}
\end{figure}

Figure \ref{fig:f1_msa_pal_aug_t_f} shows that training data augmentation does not affect the performance on the clean MSA and PAL test sets. On the contrary, at the lowest resource settings the augmented model out-performed the non-augmented on MSA and PAL by 0.43\% and 0.58\%, respectively. At 50\% and 100\% settings, both the augmented and non-augmented models' performance converge on MSA and PAL.

\section{Conclusion}
\label{sec:conclusion}
In this paper, we presented the ArBanking77 dataset, consisting of queries in both MSA and Palestinian dialects in the banking domain. As far as we know, ArBanking77 is the first Arabic intent detection dataset in the banking domain. The dataset contains 31,404 queries and 77 intents. The data was then used to fine-tune a BERT-based model for the intent detection task, resulting in an F1-score of 0.9209 for MSA and 0.8995 for PAL. We also simulated low-resource settings and found that the model is robust and with only 20\% of the data, model performance on PAL and MSA dropped by only 6.32\% and 4.51\%, respectively. We noted that training data augmentation does not negatively affect the model performance on the clean MSA and PAL test sets. In fact, at the lowest resource settings (20\%) the augmented model out-performed the non-augmented model on both MSA and PAL. 

We performed additional data augmentation to simulate errors, misspellings, and other mistakes that are common in real NLP systems. We observed the accuracy on PAL-\textit{sim\textsubscript{s}} suffers greatly when the model is trained on 20\% of the non-augmented data. Augmenting the training data closes the performance gap on PAL-\textit{sim\textsubscript{s}} by about 5\%. This indicates that BERT is susceptible to some errors, especially in dialectal Arabic which has less consistent orthography than MSA. It is also noticeable that the relative drop in accuracy between the 20\% and 50\% training sets is much larger than 50\% and 100\% case. This implies that the negative effect of the introduced errors in the dialectical Arabic is inversely proportional to the amount of data used in the train set. Finally, based on the low performance using zero-shot learning on MSA and PAL and a slight lexical overlap between them, we concluded that there is an urgent need to annotate MSA and dialectical Arabic.

\section{Limitations}
\label{sec:limitations}
Our dataset is limited to MSA and Palestinian dialect and covers only 77 intents. Applying our models and data to dialects others than MSA and PAL may not yield accurate intents. Furthermore, our data covers intents that are commonly found in traditional banking. Additional intents may need to be studied from non-traditional banking such as Islamic banks. We plan to extend our dataset to cover more Arabic dialects and obtain data from non-traditional banking institutions in the Arab region to better understand the difference in intents compared to the traditional banking. Moreover, we want to explore natural language understanding in the banking domain by combining named entity recognition with intent detection. 

We can further improve model performance by adding additional auxiliary loss functions such as contrastive loss, which will help align the token representations between the MSA and PAL queries. Furthermore, due to data limitation, the models trained on the data, including Banking77, perform intent classification using a single utterance. In practice, the query has a context, preceding utterances, that can provide important signal to the model, which may lead to better performance.

\section*{Acknowledgements}
This research is partially funded by the Palestinian Higher Council for Innovation and Excellence. We would like to thank Taymaa Hammouda for the technical support. The authors also acknowledge the great efforts of many students who helped in the annotation process, especially Rania Shahwan, Dalal Bawatneh, Malak Elsheik, Maissan Qadi, Marah Beirat, Sara Shaabna, Hiba Qasrawee, Manar Jawabreh, Mohammad AbuBader, Tuqa Qurt, Rahaf Bakeer, Marah Qoud, Nirmeen Al-Sheikh, Ameer Eleyan, Sondus Ilawi, Fatima Kusbeh, Ameena Jadallah, Eyab Ghifari, Lina Salameh, Shorouq Zaid, Mariam Abdelqader, Noor Momani, Yazan Assaf, Hala Hamza, Sondus Majdobeh and Sawsan Mohamad.

\bibliography{custom,MyReferences}
\bibliographystyle{acl_natbib}

\appendix

\section{Error Simulation Types}
\label{app:err_sim}

\subsection{Common Errors (\textit{sim\textsubscript{c}})}
\textit{sim\textsubscript{c}} are common spelling errors and word variations that people often make in real-life, which we derive from a lexicon. In a previous work, we developed a lexicon that contains a list of base forms, and the lexical variants (mostly colloquial terms) of each base form. The lexicon curation process started by collecting data from social media sites, chatbots and call centers audio recordings, which were transcribed manually. For each lexical variant, colloquial term and misspelling, the goal was to find its corresponding base form. Hence, a base form in the lexicon can have more than one lexical variant. The lexicon contains 12,111 base forms. To simulate these errors in our intent detection task, for each query, we randomly selected one to two words that have a matching base form in the lexicon, and for each base form we randomly selected one of its lexical variants. Because these errors are not simulated and are mostly colloquial variants collected from real content, we injected this type of error into the test set only, which will give us an insight how robust the model's performance is on such noisy data. Examples of orthographic variants are shown in Table \ref{tab:lexical_varations}. For instance, the world {\small \<شكرا>}/thanks has four variants ({\small \<ششكرا>}, {\small\<شككرا>}, {\small \<شكرااااااا>}, and {\small \<شكررا>}). 

\begin{table}[ht!]
\small
\centering
\begin{tabular}{|l|r|r||r|}
\hline
Lexical Term & Lexical Variants \\ \hline
\<شكرا>\  & \<شكرااااااا>\\\ 
  & \<ششكرا>\\\
& \<شكررا>\ \\
& \<شككرا>\ \\ \hline
\<مثلا>\   & \<مثلاث>\ \\ 
& \<مثل>\ \\ \hline

\end{tabular}
\caption{Sample of lexicon, some words are colloquial while others are misspellings.}
\label{tab:lexical_varations}
\end{table}

\subsection{Simulated Errors (\textit{sim\textsubscript{s}})}
\textit{sim\textsubscript{s}} are errors simulated by deleting spaces between words. We applied this type of simulation on both the train and test sets. For each query we randomly deleted one or two spaces.

\subsection{Keyboard Errors (\textit{sim\textsubscript{k}})}
\textit{sim\textsubscript{k}} are errors generated by inserting or deleting a letter from a word, replacing a letter with another letter, or swapping the places of two adjacent letters. Two approaches we followed when simulating this error. Either random replacement or replacement guided by the keyboard layout of the target language. Keyboard layout guided simulation will delete/insert/replace/swap letters based on the neighboring letters on the keyboard.

\end{document}